\documentclass{article}

 \usepackage[dblblindworkshop, final]{neurips_2025}
\workshoptitle{New Perspectives in Graph Machine Learning}

\usepackage[utf8]{inputenc} % allow utf-8 input
\usepackage[T1]{fontenc}    % use 8-bit T1 fonts
\usepackage{hyperref}       % hyperlinks
\usepackage{url}            % simple URL typesetting
\usepackage{booktabs}       % professional-quality tables
\usepackage{amsfonts}       % blackboard math symbols
\usepackage{nicefrac}       % compact symbols for 1/2, etc.
\usepackage{microtype}      % microtypography
\usepackage{xcolor}         % colors
\usepackage{amsmath}
\usepackage{multirow}
\usepackage{graphicx}

\title{KAN-GCN: Combining Kolmogorov–Arnold Network with Graph Convolution Network for an Accurate Ice Sheet Emulator}

\author{%
  Zesheng Liu \\
  Department of Computer Science and Engineering\\
  Lehigh University\\
  Bethlehem, PA 18015 \\
  \texttt{zel220@lehigh.edu} \\
  \And
  YoungHyun Koo \\
  National Snow and Ice Data Center\\
  Cooperative Institute for Research in Environmental Sciences\\
  University of Colorado Boulder\\
  Boulder, CO 80309\\
  \texttt{younghyun.koo@colorado.edu}
  \And
  Maryam Rahnemoonfar \\
  Department of Computer Science and Engineering\\
  Department of Civil and Environmental Engineering\\
  Lehigh University\\
  Bethlehem, PA 18015 \\
  \texttt{maryam@lehigh.edu} \\
}

\begin{document}

\maketitle

\begin{abstract}
We introduce KAN–GCN, a fast, accurate emulator for ice-sheet modeling that places a Kolmogorov–Arnold Network (KAN) as a feature-wise calibrator before graph convolution networks (GCNs). The KAN front end applies learnable 1-D warps and a linear mix, improving feature conditioning and nonlinear encoding without increasing message-passing depth. We employ this architecture to improve the performance of emulators for numerical ice sheet models. Our emulator is trained and tested using 36 melting-rate simulations with 3 mesh-size settings in the Pine Island Glacier, Antarctica. Across 2–5 layer architectures, KAN–GCN reproduces or outperforms the accuracy of pure GCN and MLP–GCN baseline. Despite a small parameter overhead, KAN–GCN improves inference throughput on coarser meshes by replacing one edge-wise message-passing layer with a node-wise transform; only the finest mesh shows a modest cost. Overall, KAN-first designs offer a favorable accuracy–efficiency trade-off for large transient scenario sweeps.
\end{abstract}

\section{Introduction}
Rapid climate-driven mass loss from the Greenland and Antarctic ice sheets is now a major contributor to sea-level rise, underscoring the need to accurately model ice-sheet dynamics\cite{Otosaka2023, Forsberg2017, zwally2011, Rignot2011}. Although various numerical models can reveal how ice sheets interact with climate change by solving coupled, nonlinear PDEs grounded in glaciological physics, they are computationally intensive. To improve efficiency, machine-learning (ML) emulators have been widely explored.  Once trained, they provide fast and low-cost inference by exploiting modern GPUs, which enables rapid scenario analysis while complementing physics-based modeling.

In particular, to emulate the unstructured meshes of numerical models, Koo et al.\cite{Koo_Rahnemoonfar_2025, Koo_Helheim} proposed to employ graph convolutional networks (GCNs) as fast GPU-accelerated emulators. Unlike traditional convolutional neural networks, which are tailored to data on regular Euclidean grids, GCNs naturally operate on irregular, non-Euclidean structures, making them particularly well-suited to represent complex ice-sheet boundaries and mesh-based simulation outputs from traditional numerical models. Despite their success as the first attempt to use GCN architecture for ice sheet application, their GCN emulator still exhibited substantial errors, particularly when emulating ice velocity.

\begin{figure*}[ht]
\begin{center}
\centerline{\includegraphics[width=0.7\textwidth]{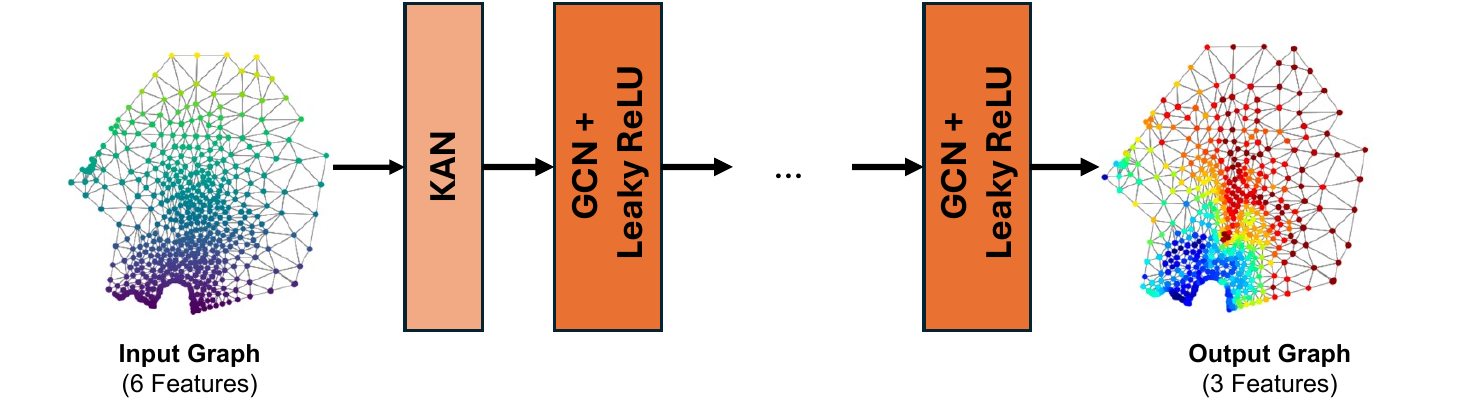}}
\caption{Schematic illustration of our proposed KAN-GCN emulator.}
\label{fig:arch}
\end{center}
\end{figure*}

In this work, we propose KAN-GCN, a hybrid emulator that places a Kolmogorov–Arnold Network(KAN)\cite{liu2025kankolmogorovarnoldnetworks} encoder in front of a stack of graph convolutional layers, aiming to further improve the accuracy of deep learning emulators for Pine Island Glacier (PIG), Antarctica. Figure~\ref{fig:arch} shows the architecture of our proposed KAN-GCN emulator. The KAN performs per-feature 1-D nonlinear expansions and mixing, learning interpretable response curves for key physical drivers (e.g., melt rate, surface mass balance, temperature), while the GCN handles spatial message passing on the mesh to propagate local interactions.

Through our experiments, we found that this KAN-GCN design can offer three main advantages compared with pure GCN-based emulators: 
\begin{itemize}
    \item Higher accuracy, especially for velocity emulation, by capturing strong feature-wise nonlinearities before spatial aggregation
    \item Improved conditioning and sample efficiency via feature calibration at the input, especially on coarser meshes that contain fewer nodes and edges
    \item Better interpretability and portability, since KAN’s univariate functions reveal variable effects and the GCN remains compatible with irregular meshes and GPU-efficient inference
\end{itemize}

% \section{Related Work}
\section{Methodology}

\subsection{Kolmogorov–Arnold Network}
We begin with the Kolmogorov-Arnold Representation Theorem\cite{Kolmogorov1957}. As proposed by Vladimir Arnold and Andrey Kolmogorov, any multivariate continuous function on a bounded domain can be represented as a finite superposition of continuous univariate functions and addition. More specifically, for any continuous $f:[0,1]^n\to\mathbb{R}$, there exist continuous univariate functions
$\phi_{q,p}:[0,1]\to\mathbb{R}$ and
$\Phi_q:\mathbb{R}\to\mathbb{R}$ such that
\begin{equation}
\label{eq:kart}
f(x_1,\ldots,x_n)\;=\;\sum_{q=1}^{2n+1}\,\Phi_q\!\Big(\,\sum_{p=1}^{n}\phi_{q,p}(x_p)\Big).
\end{equation}

Building on this idea, Liu et al.\cite{liu2025kankolmogorovarnoldnetworks} proposed Kolmogorov--Arnold Networks(KANs) that generalizes Eq.~(\ref{eq:kart}) to arbitrary widths and depths. They parameterize the univariate functions $\phi$ as B-spline curves, with learnable coefficients of local B-spline basis functions\cite{liu2025kankolmogorovarnoldnetworks}. Given an input $\,\mathbf{h}\in\mathbb{R}^{n_{\text{in}}}\,$, a KAN layer produces $\,\mathbf{z}\in\mathbb{R}^{n_{\text{out}}}\,$ via a matrix of 1D functions:
\begin{equation}
\label{eq:kan-layer}
z_j \;=\; \sum_{i=1}^{n_{\text{in}}} \,\phi_{j i}(h_i), 
\qquad j=1,\ldots,n_{\text{out}},
\end{equation}
where each edge function $\phi_{j i}$ is parameterized by a B-spline basis together with a residual (identity) term:
\begin{equation}
\label{eq:kan-spline}
\phi_{j i}(u)
=w^{(b)}_{j i} b(u) + w^{(s)}_{j i}\text{spline}(u)
\end{equation}
where $b(u)=\text{silu}(u)=u/(1+e^{-u})$, $\text{spline}(u)$ is parameterized as a linear combination of B-splines such that $\text{spline}(u)=\sum_ic_iB_i(u)$, with trainable parameters $c_i$. Eqs.~(\ref{eq:kan-layer})-(\ref{eq:kan-spline}) capture KAN’s core design: nodes sum incoming activations, while edges carry learnable univariate activations implemented as splines; the residual term stabilizes optimization \cite{liu2025kankolmogorovarnoldnetworks}. 

In order to reduce the training time and improve efficiency, we use the FastKAN setting\cite{li2024kolmogorovarnold}, which replaces B-splines with Gaussian radial basis functions (RBFs), defined as:
\begin{equation}
\label{eq:fastkan}
B_i(u) \approx 
\exp\Big(-\,\frac{(u-\mu_g)^2}{2\sigma_g^2}\Big),
\end{equation}
where $\mu_g$ are fixed centers (e.g., uniform spaced) and $\sigma_g$ are fixed, shared bandwidths determined by the grid spacing that control the spread of the RBFs. This substitution preserves the formulation in Eq.~(\ref{eq:kan-layer}) while yielding simpler, GPU-friendly computations with comparable accuracy \cite{li2024kolmogorovarnold}. In our model, we use Eq.~(\ref{eq:fastkan}) as the node-wise encoder before graph propagation via graph neural networks.

\subsection{Graph Convolution Network}
After node-wise encoding, we propagate information over a fixed spatial graph with a stack of Graph Convolution Networks(GCNs)\cite{kipf2017semi}. Inspired by first-order approximation of localized spectral filters on graphs\cite{Defferrard2016,HAMMOND2011129}, the layer-wise update rule of GCN can be defined as:
\begin{equation}
\label{eq:gcn}
\mathbf{H}^{(\ell+1)}=
\sigma\left(\tilde{\mathbf{D}}^{-\frac{1}{2}}\tilde{\mathbf{A}}\tilde{\mathbf{D}}^{-\frac{1}{2}}
\mathbf{H}^{(\ell)}\mathbf{W}^{(\ell)}\right)
\end{equation}
where $\tilde{\mathbf{A}}=A+I_N$ is the adjacent matrix of the undirected graph with added self-connections, $I_N$ is a identity matrix, $\tilde{D}_{ii} = \sum_{j} \tilde{A}_{ij}$, $W$ is the layer-specific trainable weight matrix, $H^{(l)} \in \mathbb{R}^{N \times D}$ is the matrix of activations in the $l^{th}$ layer, $H^{(0)}=X$, and $\sigma(\cdot)$ is the activation function. 

In our proposed KAN–GCN emulator, we use Leaky ReLU with a negative slope of 0.01 as the activation function. The GCN stack is initialized with $H^{(0)}$, the KAN-encoded node features, and all layers operate on a shared graph topology. By decoupling feature-wise nonlinear encoding (KAN) from local spatial aggregation (GCN), the design improves training stability and sample efficiency.

\subsection{Training Objective Refinements}
\label{other_design}
\textbf{Task and Target Reformulation:} In prior work, Koo et al.~\cite{Koo_Rahnemoonfar_2025} trained an emulator that takes the initial physical state together with a time index $t$ and directly predicts the ice thickness and velocities at timestep $t$. In contrast, we reformulate the task as learning a one-step residual update. Let $v_t=[V_{x, t}, V_{y, t}]$ and $H_t$ denote the velocity and thickness at time $t$, we then define the residual values to be $\Delta v_t = v_t-v_{t-1}$ and $\Delta H_t=H_t-H_{t-1}$. Our proposed KAN-GCN emulator is trained to predict $\widehat{\Delta v_t}$ and $\widehat{\Delta H_t}$ from $v_{t-1}$, $H_{t-1}$, together with melting rate, surface mass balance and time. After prediction, we can do the following reconstruction:
\begin{align}
\hat{v}_t &= v_{t-1} + \widehat{\Delta v_t} \\
\hat{H}_t &= H_{t-1} + \widehat{\Delta H_t}
\end{align}
This residual formulation shortens the temporal gap and focuses learning on incremental dynamics rather than the absolute value. It yields better conditioning and calibration because that the new targets are near zero-mean deltas. This design also aligns with finite-difference update rules in physical solver, so multi-step rollout is a recursive application of the same learned operator while preserving the emulator framing as an accurate one-step transition.

\textbf{Feature-wise Loss Re-weighting:} Instead of a single mean-squared error (MSE) loss over the concatenated target, we separate velocity and thickness and apply feature-specific weights to reflect scale, noise, and priority. During training, we compute a weighted loss on the reconstructed $\hat{v}_t$ and $\hat{H}_t$, rather than directly on predicted delta values, as shown in Eq.~(\ref{loss}). Because mean-squared error is translation-invariant, this is exactly equivalent to a delta-space loss and the whole training process remains fully consistent with the residual formulation. To balance differences learning priority and noise level, we sum separate loss terms for velocity and thickness with coefficients $\lambda_v$ and $\lambda_h$. In addition, before computing the weighted MSE loss we multiply residuals by a global scale factor $s$, which simply scales the objective by $s^2$ without changing the minimizer; this enlarges gradients for small updates and improves numerical conditioning and the responsiveness of learning rate schedulers.

% \begin{equation}
% \label{loss}
% \mathcal{L}(s)
% = s^{2}\!\left[
% \lambda_v \big\| \widehat{\Delta v_t} - \Delta v_t \big\|_2^2
% + \lambda_h \big\| \widehat{\Delta H_t} - \Delta H_t \big\|_2^2
% \right]
% = s^{2}\!\left[
% \lambda_v \big\| \hat{v}_t - v_t \big\|_2^2
% + \lambda_h \big\| \hat{H}_t - H_t \big\|_2^2
% \right],
% \end{equation}

\begin{equation}\label{loss}
\begin{aligned}
\mathcal{L}(s)
&= s^{2}\!\left[
\lambda_v \big\| \widehat{\Delta v_t} - \Delta v_t \big\|_2^2
+ \lambda_h \big\| \widehat{\Delta H_t} - \Delta H_t \big\|_2^2
\right] \\
&= s^{2}\!\left[
\lambda_v \big\| \hat{v}_t - v_t \big\|_2^2
+ \lambda_h \big\| \hat{H}_t - H_t \big\|_2^2
\right].
\end{aligned}
\end{equation}

\section{Experiment Setup}
\subsection{Data Preparation}
In order to generate datasets for our proposed KAN-GCN emulator, we implement transient simulations on the Ice-sheet and Sea-level System Model (ISSM). We acquire numerical simulations of the 20-year evolution of ice thickness and velocity in the Pine Island Glacier, Antarctica, by adapting the sensitivity experiments\cite{Larour_2012,Seroussi_2014} governed by Shelfy-Stream Approximation\cite{Morland_1987,MacAyeal_1989}. Given that this glacier has shown the fastest ice speed and the largest contribution to the sea-level rise in Antarctica, this glacier simulation data is appropriate for testing the effectiveness of our approach in lowering the uncertainties in future sea-level prediction.

To study how our proposed deep learning emulator performs on different mesh resolutions, we run ISSM transient simulations on three different mesh sizes: 2km, 5km, and 10km. Motivated by evidence that basal melt is the primary driver of mass loss at Pine Island Glacier \cite{Jacobs2011,Joughin2021}, we also explore annual basal-melt scenarios from 0 to $70\,\mathrm{m\,a^{-1}}$ in $2\,\mathrm{m\,a^{-1}}$ increments to assess their impact on ice-sheet dynamics. In summary, we ran the 20-year ISSM transient simulations on Pine Island Glacier 108 times (3 mesh sizes, each with 36 melting rates).

We construct our graph structure based on the raw mesh of ISSM simulations: the nodes and edges in each triangular mesh of ISSM will become the nodes and edges in our graph dataset. 
Instead of using all the features from ISSM transient simulation as inputs to the emulator~\cite{Koo_Rahnemoonfar_2025}, 
we use a more compact input representation based on our experimental preference. In particular, we exclude the velocity magnitude channel and keep only the two velocity components $V_x$ and $V_y$. We also choose to keep thickness and not include surface or base elevations, and we do not include the ice-floating field. After this pruning, our emulator uses six input channels, melting rate, surface mass balance, time, $V_x$, $V_y$, and thickness, and is trained to predict $V_x$, $V_y$, and thickness at next timestep. Therefore, each node contains 6 input node features (melting rate, time $t-1$, surface mass balance at time $t-1$, x-component and y-component velocity at time $t-1$, thickness at time $t-1$) and 3 output features(x-component and y-component velocity at time $t$, thickness at time $t$). To ensure the stability of the training process, we follow the same pre-processing procedure as Koo et al.~\cite{Koo_Rahnemoonfar_2025} that normalize all the input and output features to the range of $[-1, 1]$ using the nominal maximum and minimum values that each feature may have.

%\subsection{Feature Curation}

%Compared with the GCN emulator proposed by Koo et al.\cite{Koo_Rahnemoonfar_2025}, we adopt a lightweight feature curation to stabilize training and improve performance. Specifically, we remove the summed speed (velocity magnitude) channel because it is fully determined by the two velocity components $V_x$ and $V_y$ and adds no new information. We omit surface and base elevations once thickness is included, since those three variables are linearly linked and thickness is the more compact, physically meaningful representation for our task. We also drop the ice-floating field, as it can be sensitive to meshing and boundary choices, and may be determined by other features. After pruning, our emulator will have six input channels, including melting rate, surface mass balance, time, $V_x$, $V_y$ and thickness, aiming to predict $V_x$, $V_y$ and thickness at a different time step. We follow the same pre-processing methods by Koo et al.\cite{Koo_Rahnemoonfar_2025} that scale each feature to the range of $(0, 1)$ using scale factors.

\subsection{Training Details}
For the total of 25,812 graphs (239 residual values times 3 mesh grid sizes times 36 melting rates), we divide them into training, validation and testing datasets based on the melting rate values. Specifically, graphs with melting rate of 0, 20, 40 and 60 $m\cdot a^{-1}$ are used for validation, graphs with melting rate of 10, 30, 50, and 70 $m\cdot a^{-1}$ are used for testing. As a result, the number of graphs in the training, validation and testing datasets is 20076, 2868, and 2868, respectively. By stratifying experiments across melt-rate regimes, we test whether our GNN emulators generalize to out-of-distribution melting conditions.

In order to demonstrate the effectiveness of our proposed KAN-GCN emulator, we compared it against a purely GCN-based emulator and an MLP-GCN-based emulator. In the MLP–GCN baseline, the KAN feature encoder is replaced by a standard node-wise MLP that maps raw node features to the same hidden dimension as KAN; the downstream GCN stack, graph connectivity, depth, and training hyperparameters are kept identical. In the pure GCN baseline, the KAN encoder is replaced by a GCN layer that performs the input projection to the hidden width; total depth is matched by inserting this projection layer into the stack, with all other settings held fixed. We evaluated architectures with depths ranging from 2 to 5 layers to quantify the effect of using a KAN as the first layer across different model complexities. All the networks are trained using the same GPU instance\cite{Jetstream2_2, Jetstream2_1}, which is composed of an AMD EPYC Milan processor, 120 GB of system RAM, and a single NVIDIA A100 GPU with 40 GB of VRAM. Adam\cite{kingma2017adammethodstochasticoptimization} is used as the optimizer. For all baselines that do not use the re-weighted loss function (Equation~\ref{loss}), we train them with a learning rate of 0.01 and use an exponential learning rate scheduler with gamma equal to 0.99, while for other advanced emulators, we use a learning rate of 0.0005, together with the cosine anneal scheduler\cite{loshchilov2017sgdrstochasticgradientdescent}. To ensure the convergence of each emulator, we train all of them for 500 epochs. All the deep learning emulators are implemented with the Deep Graph Library (DGL)\cite{wang2019dgl} and PyTorch modules.

 % We also include the emulator proposed by Koo et al.\cite{Koo_Rahnemoonfar_2025} and its two variants that only include the training objective refinements as baseline models.
\section{Result and Analysis}

\subsection{Accuracy}
We evaluate the accuracy of our proposed KAN-GCN emulator by calculating the root mean squared error(RMSE) between the emulator prediction $y$ and ISSM simulation results $\hat{y}$, defined as:
\begin{equation}
    \label{rmse}
    RMSE(\hat{y}, y)=\sqrt{\frac{1}{N}\sum_{i=1}^N(\hat{y_i}-y_i)^2}
\end{equation}

\begin{table}[ht]
    \centering
    \caption{RMSE error on thickness prediction with different mesh grid sizes}
    \label{tab:rmse_thickness}
\scalebox{0.72}{
\begin{tabular}{ccccc}
\toprule
\multirow{2}{*}{Method}               & \multirow{2}{*}{Network Architecture} & \multicolumn{3}{c}{RMSE error on Thickness H} \\ \cline{3-5} 
                                      &                                       & Mesh 2000             & Mesh 5000             & Meth 10000             \\ \midrule
Koo et al.\cite{Koo_Rahnemoonfar_2025}                            & 5GCN                                  & 12.9629               & 13.9422               & 14.9061                \\
$t-1 \rightarrow t$                   & 5GCN                                  & 9.0505                & 10.8579               & 11.7124                \\
$t-1 \rightarrow \Delta t$ + Feature Curation                   & 5GCN                                  & 0.9212                & 0.9467                & 0.9616                 \\ \midrule
$t-1 \rightarrow \Delta t$ + Feature Curation + Loss re-weight & 2GCN                                  & \textbf{0.4790}       & \textbf{0.4773}       & \textbf{0.5158}                 \\
$t-1 \rightarrow \Delta t$ + Feature Curation + Loss re-weight & MLP+GCN                               & 0.6980                & 0.6898                & 0.7269                 \\ 
$t-1 \rightarrow \Delta t$ + Feature Curation + Loss re-weight & KAN+GCN (Ours)                               & 0.5996                & 0.5919                & 0.6314                 \\ \midrule
$t-1 \rightarrow \Delta t$ + Feature Curation + Loss re-weight & 3GCN                                  & 0.4817                & 0.4725                & 0.5156                 \\
$t-1 \rightarrow \Delta t$ + Feature Curation + Loss re-weight & MLP+2GCN                              & 0.4930                & 0.4928                & 0.5299                 \\
$t-1 \rightarrow \Delta t$ + Feature Curation + Loss re-weight & KAN+2GCN (Ours)                             & \textbf{0.4616}       & \textbf{0.4583}       & \textbf{0.4890}                 \\ \midrule
$t-1 \rightarrow \Delta t$ + Feature Curation + Loss re-weight & 4GCN                                  & 0.4119                & 0.4188                & 0.4535                 \\
$t-1 \rightarrow \Delta t$ + Feature Curation + Loss re-weight & MLP+3GCN                              & 0.4244                & 0.4242                & 0.4573                 \\ 
$t-1 \rightarrow \Delta t$ + Feature Curation + Loss re-weight & KAN+3GCN (Ours)                              & \textbf{0.4043}       & \textbf{0.4061}       & \textbf{0.4321}                 \\ \midrule
$t-1 \rightarrow \Delta t$ + Feature Curation + Loss re-weight & 5GCN                                  & 0.3901                & 0.4046                & 0.4376                 \\
$t-1 \rightarrow \Delta t$ + Feature Curation + Loss re-weight & MLP+4GCN                              & 0.3896                      &  0.4011                     &   0.4370                     \\ 
$t-1 \rightarrow \Delta t$ + Feature Curation + Loss re-weight & KAN+4GCN (Ours)                              & \textbf{0.3850}       & \textbf{0.3923}    & \textbf{0.4245}                        \\ \hline
\end{tabular}
}
\end{table}

\begin{table}[ht]
    \centering
    \caption{RMSE error on velocity prediction with different mesh grid sizes}
    \label{tab:rmse_velocity}
\scalebox{0.72}{
\begin{tabular}{ccccc}
\toprule
\multirow{2}{*}{Method}               & \multirow{2}{*}{Network Architecture} & \multicolumn{3}{c}{RMSE error on velocity $V_x,V_y$} \\ \cline{3-5} 
                                      &                                       & Mesh 2000             & Mesh 5000             & Meth 10000             \\ \midrule
Koo et al.\cite{Koo_Rahnemoonfar_2025}                            & 5GCN                                  & 56.0270               & 50.0802               & 52.3956                \\
$t-1 \rightarrow t$                   & 5GCN                                  & 20.7602                & 25.3250               & 27.3886                \\
$t-1 \rightarrow \Delta t$ + Feature Curation                   & 5GCN                                  & 4.7475                & 4.5094                & 4.0734                 \\ \midrule
$t-1 \rightarrow \Delta t$ + Feature Curation + Loss re-weight & 2GCN                                  & \textbf{2.7285}       & \textbf{2.7723}       & \textbf{2.5860}                 \\
$t-1 \rightarrow \Delta t$ + Feature Curation + Loss re-weight & MLP+GCN                               & 3.9099                & 3.7869                & 3.3624                 \\ 
$t-1 \rightarrow \Delta t$ + Feature Curation + Loss re-weight & KAN+GCN (Ours)                               & 3.3624                & 3.3774                & 3.1260                 \\  \midrule
$t-1 \rightarrow \Delta t$ + Feature Curation + Loss re-weight & 3GCN                                  & 2.6371                & 2.7371                & 2.5394                 \\
$t-1 \rightarrow \Delta t$ + Feature Curation + Loss re-weight & MLP+2GCN                              & 2.7663                & 2.7558                & 2.6222                 \\
$t-1 \rightarrow \Delta t$ + Feature Curation + Loss re-weight & KAN+2GCN (Ours)                             & \textbf{2.3612}       & \textbf{2.4314}       & \textbf{2.2767}                 \\ \midrule
$t-1 \rightarrow \Delta t$ + Feature Curation + Loss re-weight & 4GCN                                  & 2.1913                & 2.3661                & 2.1485                 \\
$t-1 \rightarrow \Delta t$ + Feature Curation + Loss re-weight & MLP+3GCN                              & 2.2541                & 2.3731                & 2.1851                 \\ 
$t-1 \rightarrow \Delta t$ + Feature Curation + Loss re-weight & KAN+3GCN (Ours)                              & \textbf{2.0407}       & \textbf{2.1897}       & \textbf{2.0159}        \\ \midrule
$t-1 \rightarrow \Delta t$ + Feature Curation + Loss re-weight & 5GCN                                  & 2.0763                & 2.2514                & 2.0459                 \\
$t-1 \rightarrow \Delta t$ + Feature Curation + Loss re-weight & MLP+4GCN                              & 2.0452                      &  2.2094                     &  \textbf{1.9696}                     \\
$t-1 \rightarrow \Delta t$ + Feature Curation + Loss re-weight & KAN+4GCN (Ours)                              & \textbf{1.9212}       & \textbf{2.0364}    & 1.9789                       \\ \hline
\end{tabular}
}
\end{table}

Table~\ref{tab:rmse_thickness} and Table~\ref{tab:rmse_velocity} show the experiment results. Compared with baseline pure-GCN emulator proposed by Koo et al.\cite{Koo_Rahnemoonfar_2025}, applying feature curation and training objective refinements can effectively imporve the overall accuracy. Across depths, KAN–GCN is consistently more accurate than pure GCN for 3–5 layers but slightly worse at 2 layers. This pattern is consistent with the roles of aggregation and encoding: in a 2-layer stack, replacing one GCN layer with a KAN front end reduces message-passing depth (one hop instead of two) and thus limits spatial aggregation, offsetting the benefit of KAN’s feature-wise nonlinear encoding; once the stack reaches 3–5 layers, message passing is ample, and the KAN front end improves conditioning and mitigates early oversmoothing, yielding higher accuracy than a pure GCN.

For the thickness prediction that tends to be a more easier task, all the emulators can achieve high accuracy while our proposed KAN-GCN emulator marginally performs better than pure GCNs or MLP-GCN emulator. On the velocity prediction, KAN-GCN emulator significantly outperforms other emulators, with a corner case existing for the 5-layer architecture and 10 km mesh size. We attribute this to the fact that velocity is more nonlinear and anisotropic compared to thickness, so KAN’s adaptive univariate bases capture feature-specific responses more effectively. For the exception, we think in that scenario with fewer nodes and edges, together with a deep aggregation that stacks four GCN layer, the pattern is simpler so the lighter MLP front end regularizes better and avoids the mild overfitting that a higher-capacity KAN can induce. Overall, these results suggest using KAN as the first layer is most beneficial once sufficient message-passing depth is available and for targets with stronger nonlinearity (e.g., velocity), which benefit from feature-wise nonlinear encoding and improved conditioning via per-feature calibration (learnable 1-D warps) that preconditions inputs before spatial aggregation without increasing message-passing depth.

\subsection{Computational Performance}
To evaluate the computational performance of our KAN-GCN emulator with 2-, 3-, 4-, and 5-layer architecture, we record the elapsed time that each emulator generates the final 20-year transient simulation results with 36 melting rates on the same GPU.

\begin{table}[ht]
    \centering
    \caption{Number of parameters for each emulator and total elapsed time(in seconds) on GPU for producing 20-year transient simulation with 36 melting rates}
    \label{tab:time}
\scalebox{0.7}{
\begin{tabular}{cccccc}
\hline
\multirow{2}{*}{Method}               & \multirow{2}{*}{Network Architecture} & \multirow{2}{*}{\# of parameters} & \multicolumn{3}{c}{Total Computation Time}  \\ \cline{4-6} 
                                      &                                       &                                             & Mesh 2000 & Mesh 5000 & Mesh 10000 \\ \toprule
$t-1 \rightarrow \Delta t$ + Feature Curation + Loss re-weight & 2GCN                                  & 17795                                       & \textbf{3.3626}    & 2.4677    & 2.4010     \\
$t-1 \rightarrow \Delta t$ + Feature Curation + Loss re-weight & KAN+GCN                               & 23959                                       & 4.0236    & \textbf{2.2114}    & \textbf{2.1583}     \\
$t-1 \rightarrow \Delta t$ + Feature Curation + Loss re-weight & MLP+GCN                               & 17795                                       & 3.5106    & 2.3097    & 2.1614           \\ \midrule
$t-1 \rightarrow \Delta t$ + Feature Curation + Loss re-weight & 3GCN                                  & 34307                                       & 4.2909    & 3.3024    & 3.5949     \\
$t-1 \rightarrow \Delta t$ + Feature Curation + Loss re-weight & KAN+2GCN                              & 40471                                       & 4.3923    & 2.8601    & \textbf{2.8010}     \\
$t-1 \rightarrow \Delta t$ + Feature Curation + Loss re-weight & MLP+2GCN                              & 34307                                       & \textbf{3.6791}    & \textbf{2.8193}    & 2.8612     \\ \midrule
$t-1 \rightarrow \Delta t$ + Feature Curation + Loss re-weight & 4GCN                                  & 50819                                       & 5.0967    & 4.4377    & 4.0523     \\
$t-1 \rightarrow \Delta t$ + Feature Curation + Loss re-weight & KAN+3GCN                              & 56983                                       & 5.4000    & \textbf{3.6947}    & 3.447      \\
$t-1 \rightarrow \Delta t$ + Feature Curation + Loss re-weight & MLP+3GCN                              & 50819                                       & \textbf{4.7478}    & 3.7049    & \textbf{3.3473}     \\ \midrule
$t-1 \rightarrow \Delta t$ + Feature Curation + Loss re-weight & 5GCN                                  & 67331                                       & 6.3762    & 5.4576    & 5.7064     \\
$t-1 \rightarrow \Delta t$ + Feature Curation + Loss re-weight & KAN+4GCN                              & 73495                                       & 6.8269    & 4.6632    & \textbf{4.3957}     \\
$t-1 \rightarrow \Delta t$ + Feature Curation + Loss re-weight & MLP+4GCN                              & 67331                              & \textbf{5.4023}    & \textbf{4.5471}    & 4.5732     \\ \bottomrule
\end{tabular}
}
\end{table}

Table~\ref{tab:time} shows the total number of trainable parameters and total elapsed time of each emulator. From the experiment results, we notice that although using KAN as the first layer adds about 6.2k parameters, it does not slow inference; on the contrary, KAN–GCN is consistently faster than same-depth pure GCNs on larger (coarser) meshes that contain fewer nodes and edges. We attribute this to swapping an edge-heavy message-passing layer for a dense, node-wise KAN transform, which cuts launch/sync overhead that dominates on coarser (smaller) graphs. On the finest mesh (Mesh 2000, with the most nodes and edges), the KAN layer’s per-node spline or basis evaluations are not fully amortized, producing a modest overhead. Overall, replacing the first GCN layer with a KAN layer preserves parameter efficiency and improves inference throughput on coarse meshes, which is advantageous for large transient-scenario sweeps.

\section{Conclusion}

In this paper, we introduce KAN-GCN, an accurate and efficient emulator for ice-sheet modeling. We combine Kolmogorov--Arnold Network that provides per-feature nonlinear encoding and graph convolution for spatial coupling. Additionally, we also introduce lightweight feature curation, reformulate the target predictions of our emulator, and applied a feature-wise weigthed loss function. We compare KAN–GCN against pure GGN and MLP–GGN baselines under 2–5 layers. Placing a KAN first is most beneficial once depth $\geq$3, acting as a per-feature calibrator that improves conditioning and sample efficiency. Despite a small parameter overhead, inference efficiency is maintained or improved because a node-wise KAN replaces one edge-wise message-passing layer. We observe faster runtimes on coarser meshes and only a modest cost on the finest mesh. Accuracy gains are modest for thickness but pronounced for velocity, with one exception at 5 layers on the 10 km mesh where a lighter MLP front end regularizes better.

\bibliography{reference}
\bibliographystyle{abbrv}

\end{document}